\documentclass[preprint,12pt]{elsarticle}

\usepackage{booktabs}
\usepackage[utf8]{inputenc}
\usepackage[T1]{fontenc}
\usepackage{lmodern}
\usepackage{graphicx}
\usepackage[figurename=Fig.,labelfont=bf,labelsep=period]{caption}
\usepackage{subcaption}
\usepackage{amsmath}
\usepackage{amssymb}
\usepackage{newtxtext,newtxmath}
\usepackage[colorlinks=true,citecolor=black,linkcolor=black]{hyperref}
\usepackage[font=normalsize]{caption}
\usepackage{svg}
\usepackage{algorithm, algorithmicx, algpseudocode}
\usepackage{subcaption,siunitx,booktabs}
\usepackage{ragged2e}

\usepackage{multirow}
\usepackage{multicol}
\usepackage{makecell}

\usepackage{amssymb}
\usepackage{amsmath}

\journal{Nuclear Physics B}

\begin{document}

\begin{frontmatter}


\title{\textit{OpenGuide}: Assistive Object Retrieval in Indoor Spaces for Individuals with Visual Impairments} 


\author[1]{Yifan Xu}
\author[2]{Qianwei Wang}
\author[3]{Vineet Kamat}
\author[4]{Carol Menassa}

\affiliation[1]{ title={Ph.D. Candidate,},
organization={Department of Civil and Environmental Engineering, University of Michigan},
            city={Ann Arbor},
            postcode={48105}, 
            state={Michigan},
            country={United States},
            }
\affiliation[2]{ title={Undergraduate Student,},
organization={College of Literature, Science, and the Arts, University of Michigan},
            city={Ann Arbor},
            postcode={48105}, 
            state={Michigan},
            country={United States},
            }
\affiliation[3]{title={Professor,},organization={Department of Civil and Environmental Engineering, University of Michigan},
            city={Ann Arbor},
            postcode={48105}, 
            state={Michigan},
            country={United States}}
\affiliation[4]{title={Professor,},organization={Department of Civil and Environmental Engineering, University of Michigan},
            city={Ann Arbor},
            postcode={48105}, 
            state={Michigan},
            country={United States}}
\begin{abstract}
Indoor built environments like homes and offices often present complex and cluttered layouts that pose significant challenges for individuals who are blind or visually impaired, especially when performing tasks that involve locating and gathering multiple objects. While many existing assistive technologies focus on basic navigation or obstacle avoidance, few systems provide scalable and efficient multi-object search capabilities in real-world, partially observable settings. To address this gap, we introduce OpenGuide, an assistive mobile robot system that combines natural language understanding with vision-language foundation models (VLM), frontier-based exploration, and a Partially Observable Markov Decision Process (POMDP) planner. OpenGuide interprets open-vocabulary requests, reasons about object-scene relationships, and adaptively navigates and localizes multiple target items in novel environments. Our approach enables robust recovery from missed detections through value decay and belief-space reasoning, resulting in more effective exploration and object localization. We validate OpenGuide in simulated and real-world experiments, demonstrating substantial improvements in task success rate and search efficiency over prior methods. This work establishes a foundation for scalable, human-centered robotic assistance in assisted living environments.
\end{abstract}





\begin{keyword}
Assistive robotics, Multiple Object Search, Natural Language Interaction


\end{keyword}

\end{frontmatter}

\section{Introduction}
\label{sec:intro}

Visual impairment is used to describe any kind of vision loss, whether it is someone who cannot see at all or someone who has partial vision loss~\cite{zou2023}. According to the CDC, approximately 12 million people 40 years and over in the United States have vision impairment. This includes 1 million individuals with total blindness, and this number is predicted to more than double by 2050~\cite{CDC}. As our built environments become more visually complex and cluttered, navigating and interacting with these spaces becomes increasingly difficult for those who rely on non-visual cues~\cite{Chidiac2024}.

Among the wide range of challenges faced by people who are blind or visually impaired, navigating and interacting with indoor environments remains one of the most persistent and complex~\cite{Jeamwatthanachai2019}. Everyday activities such as cooking, dressing, or managing medications often rely on accurate spatial awareness, object recognition, and memory of one’s environment—abilities that are severely compromised without visual input and specific training~\cite{Yu2025}. Without access to reliable visual cues, individuals must depend on tactile feedback, audio prompts, or memorized layouts, which can be easily disrupted by changes in the environment. Even minor alterations, such as a misplaced item or newly introduced obstacle, can significantly increase task difficulty, leading to slower performance, confusion, and in some cases, safety risks~\cite{Lahav2008}.

Within this broader context, multi-object retrieval—the task of locating and collecting several specific objects across an indoor space—is especially challenging and frequently encountered in daily life~\cite{TroncosoAldas2020,Valipoor2024}. Tasks such as preparing a meal may involve finding utensils, specific food items, and processing tools from different a utensil from different locations in a kitchen; packing a bag may require gathering clothing, documents, and electronics from multiple rooms. For individuals with visual impairments, this process becomes cognitively demanding and labor-intensive, often involving trial-and-error searches or repeated reliance on memory. The cumulative burden of performing multi-object retrieval without visual support can lead to reduced autonomy, increased frustration, and a dependence on others for help with everyday activities~\cite{Valipoor2024}.

Currently, there are some emerging assistive technologies that address this problem. Some of the assistive technologies, such as Aira, Seeing AI, and BeMyAI help people to understand their surrounding environment, and have been commercially deployed~\cite{Kim2022}. Over the past recent years, various guidance systems have been developed to assist blind individuals with navigation~\cite{Li2016,Sato2017} and exploration~\cite{Kaniwa2024,Kayukawa2023}, providing information about their surrounding environment. Despite decades of advancement in assistive technologies, most robotic systems designed for people with visual impairments remain focused on basic mobility assistance, such as obstacle avoidance and point-to-point navigation~\cite{Chanana2017,Wong2024}. These systems typically use geometric maps and sensor data to help users avoid collisions or follow predefined paths~\cite{Chen2015,Kim2016,Murata2018}, but they lack the semantic reasoning required for goal-directed interaction with the environment. As a result, current solutions fall short when users need help with more cognitively demanding tasks, such as retrieving specific objects or understanding room-level layouts. 

In the robotics domain, there has been substantial progress in semantic environment understanding~\cite{xu2024,Werby24}, enabling robots to interpret the spatial layout and contextual information of indoor scenes. Building on this, several recent works have focused on object search in novel environments~\cite{yokoyama2024vlfm,Wandzel2019,ovamos}, where robots use vision-language models to ground user instructions and identify specific objects without prior knowledge of the environment. However, these efforts are primarily limited to single-object search tasks~\cite{Chikhalikar2024}. In real-world assistive scenarios—such as helping a person prepare a meal or pack a bag—robots must often retrieve multiple objects in sequence or in parallel~\cite{Zeng2019,ovamos}. The lack of support for multi-object search significantly limits the practicality and scalability of current systems~\cite{Shenoy2022}. Despite its importance, the development of systems capable of robust and scalable multi-object retrieval under open-vocabulary and dynamic conditions remains a research gap.

In this work, building upon our previous research in multi-object search~\cite{ovamos},we introduce a novel assistive robotic system designed to address these limitations by enabling robust multi-object retrieval in real-world indoor environments. By combining vision-language models, semantic scene understanding, and uncertainty-aware reasoning, our system interprets natural language commands, builds a semantic map of the space, and locates objects of interest with high reliability. This capability marks a significant step forward in enabling assistive robots to support not only navigation but also meaningful interaction with the environment, ultimately enhancing autonomy and quality of life for people with visual impairments.

\subsection{Statement of Contribution}

In this work, we propose OpenGuide, an assistive multi-object search framework for indoor built environment settings that can assist people with blindness and visual impairments under open-vocabulary settings. In this framework, we integrate vision-language models (VLMs), frontier-based exploration, and Partially Observable Markov Decision Process (POMDP)-based planning. The contributions are listed as follows:

\begin{itemize}
    \item We develop an open-vocabulary assistive framework that enables object retrieval in novel environments, specifically designed to support people who are blind or visually impaired.
    \item We formulate the multi-object retrieval problem as a Partially Observable Markov Decision Process (POMDP) and leverage Vision-Language Models (VLMs) to guide the robot in retrieving multiple objects efficiently with minimal travel distance.
    \item We conduct three case studies to extensively evaluate diverse indoor environments and demonstrate that our proposed framework outperforms existing baselines in both retrieval accuracy and navigation efficiency, and it can perform with consistent stability across different indoor environments.
\end{itemize}
\section{Related Work}
\label{sec:relatedwork}

\subsection{Assistive Systems for People with Visual Impairments}
Robotic guide systems have the advantage of addressing the mobility challenge for blind people with their automatic guidance capability~\cite{Thiyagarajan2022}. These systems can potentially overcome limitations of traditional aids like guide dogs, which are expensive and time-consuming to train~\cite{Fang2024}. Some of these technologies are focused on solving the mobility navigation issues for people with visual impairments by integrating computer vision and model-based navigation algorithms to help with obstacle avoidance~\cite{Galatas2011,Bruno2019}. For example, some guidance robots can navigate multiple paths, remember routes, and operate in both indoor and outdoor environments~\cite{Megalingam2018}. Lai et al.~\cite{Lai2021} designed a portable mobile robot equipped with sonar, laser scanner, depth camera, a handle with a force sensor and an elastic rope. BlindPilot, another robotic system, guided users to landmark objects more efficiently than sound-based navigation~\cite{Kayukawa2020}. Augmented Cane improved walking speed by 18\% for visually impaired participants compared to a white cane~\cite{Slade2021}, and RDog, a quadruped robot demonstrated faster and smoother navigation with fewer collisions in diverse environments~\cite{Cai2024}.

Besides the navigation problem, some technologies are focused on the exploration problem for blind people to familiarize themselves with the environment~\cite{Jain2023} or for enjoying recreational areas where exploration is essential (e.g. museums~\cite{Kayukawa2020} or shopping mall~\cite{Lacey2000}). For example, PathFinder~\cite{Kuribayashi2023} is a map-less navigation robotic system designed to guide blind users to their destination. A robotic system~\cite{Kayukawa2023} allows the users to explore by interactively setting destinations on a smartphone and by calling a museum guide to explain their surroundings. 

While recent advancements in assistive technologies have significantly improved navigation and environmental awareness for people who are blind or visually impaired, a critical gap remains in enabling efficient object retrieval at scale. Existing systems typically focus either on mobility assistance—guiding users through environments—or on environmental exploration—providing descriptions of nearby objects or landmarks~\cite{Chanana2017}. However, these capabilities fall short when it comes to supporting multi-object retrieval tasks, especially in complex and cluttered indoor spaces such as homes or offices. The lack of integrated solutions that combine semantic scene understanding, goal-driven planning, and user-intent interpretation limits the practical utility of these systems. Addressing this gap, our work developed a generalizable, open-vocabulary assistive technology that can understand natural language commands and retrieve multiple objects efficiently across diverse real-world environments.

\subsection{Multi-Object Retrieval}

In the robotics area, multi-object search (MOS) is a crucial task~\cite{gen}. Existing approaches to MOS can be broadly categorized into three types: Deep Reinforcement Learning (DRL) methods~\cite{multion,learning_long,multi_drl,multi_drl_2}, probabilistic planning methods~\cite{mos_base,mos_new,mos_system}, and foundation model--guided navigation methods. Most prior work falls into the first two categories, while foundation model-guided approaches—which leverage LLMs~\cite{esc,cat,lvmn,zson} or VLMs~\cite{vlfm,opennav}—were initially popular in single object search (SOS), and have recently been extended to MOS~\cite{finder}. 

DRL-based methods, including Deep Q Networks (DQN)~\cite{dqn}, Proximal Policy Optimization (PPO)~\cite{multion,learning_long,multi_drl_2}, and hybrid SLAM-based approaches~\cite{multi_drl}, train robots via extensive offline interactions. However, these methods often suffer from inefficient exploration due to limited semantic guidance and poor generalizability, as they require vast amounts of training data. In contrast, our open-vocabulary value map leverages foundation model cues to enhance semantic understanding and improve generalizability.

Probabilistic planning methods, typically formulated as POMDPs~\cite{mos_base,mos_new,mos_system}, manage uncertainty in object locations and sensor noise by maintaining belief states. While effective at handling observation uncertainty, these methods suffer from high computational complexity in large, cluttered environments and generally require environmental priors, limiting their adaptability to novel settings. Our method alleviates these issues by leveraging foundation model cues to simplify belief updates and reduce the action space, and by incorporating frontier-based exploration into the POMDP reward function to facilitate navigation in unfamiliar environments, thus enhancing both efficiency and robustness.

Foundation model--guided approaches utilize both natural language and visual inputs to generate rich semantic priors for navigation. By leveraging the contextual understanding of foundation models---which encompass both LLMs~\cite{esc,cat,lvmn,zson} and VLMs~\cite{vlfm,opennav}---these methods inform target localization with improved semantic cues. Early work in SOS demonstrated that combining frontier mapping with such cues leads to efficient target localization~\cite{vlfm}. For MOS, the Finder~\cite{finder} was the first to incorporate foundation model cues by constructing a multi-layer value map to guide search, achieving promising results. However, Finder~\cite{finder} struggles to recover from detector noise and occlusions because it always navigates to frontiers (i.e., the boundaries between explored and unexplored regions), which limits its ability to revisit and correct for missed detections. 

Critically, existing methods have not been tailored to the needs of people with visual impairments, who require reliable, interpretable, and efficient multi-object search in cluttered and unfamiliar indoor environments. This creates a significant research gap in developing assistive systems that can robustly navigate, interpret complex environments, and retrieve multiple objects using natural language instructions. Our framework addresses this gap by integrating foundation model cues with frontier-based exploration and POMDP planning to support robust and adaptive object search, enabling assistive robots to better serve people with visual impairments in everyday environments.

\section{METHODOLOGY}
\label{sec:methodology}

\begin{figure}
    \centering
    \includegraphics[width=\linewidth]{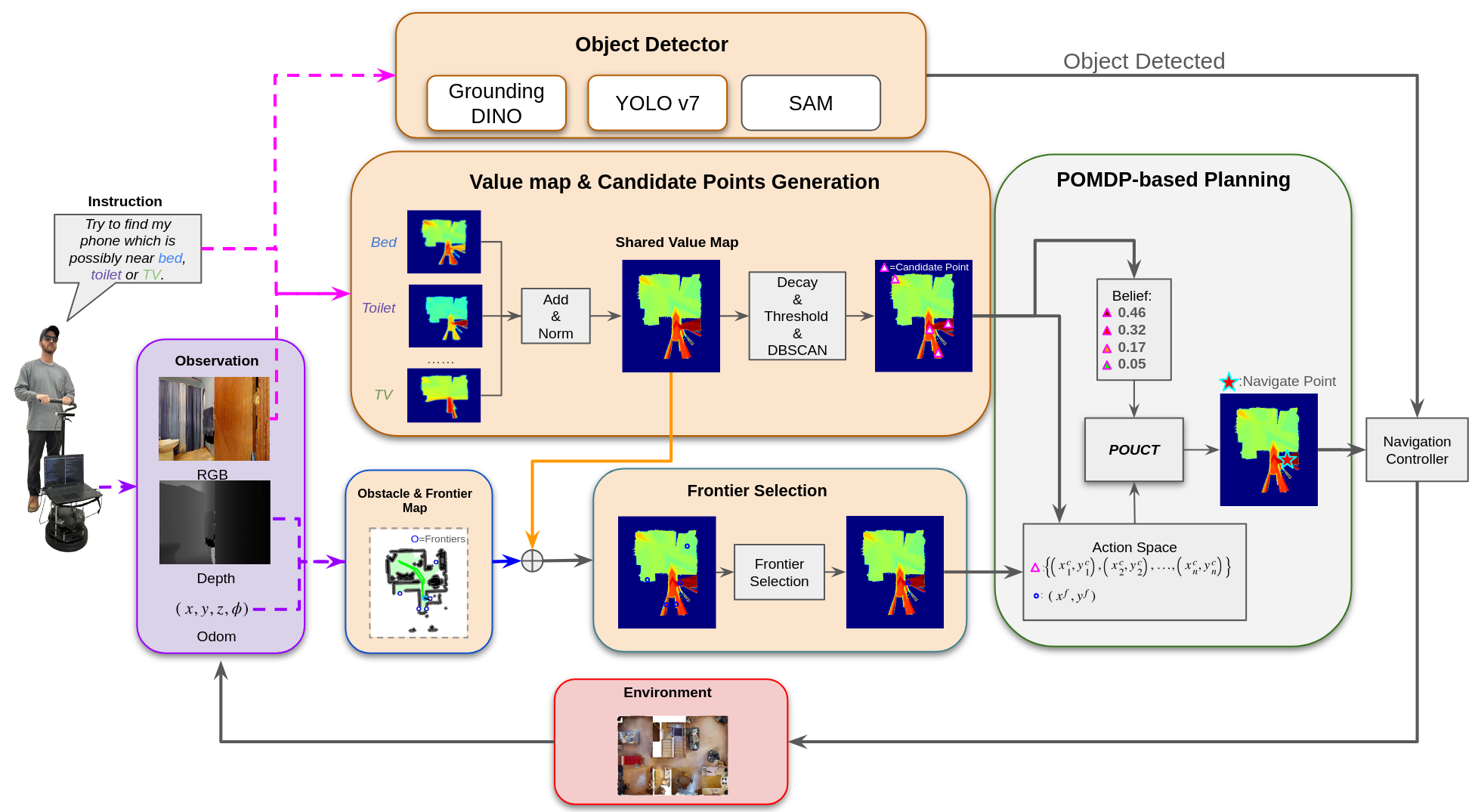}
    \caption{The OpenGuide system overview}
    \label{fig:system overview}
\end{figure}

\subsection{Problem Formulation}
The MOS problem requires a mobile robot to search for a set of \( K \) static target objects in an unknown environment. The robot's state at time \( t \) is \( \mathbf{x}_r(t) = (x, y, \phi) \in \mathbb{R}^3 \), where \( (x, y) \) is its position and \( \phi \) its orientation. The environment contains \( L \) static objects \( \mathcal{O}_{\text{env}} = \{ o_{s1}, o_{s2}, \dots, o_{sL} \} \), among which the target objects \( \mathcal{O}_{\text{tgt}} = \{ o_{t1}, \dots, o_{tK} \} \subseteq \mathcal{O}_{\text{env}} \) are to be located at unknown positions \( \mathbf{x}_{tj} \). The objective of MOS is to locate all objects in \( \mathcal{O}_{\text{tgt}} \) while minimizing the cumulative travel distance \( d = \int_0^T \| \dot{\mathbf{x}}_r(t) \| dt \), where \( T \) is the total search time.

\subsection{System Overview}

The system overview is shown in Fig.~\ref{fig:system overview}. It consists of a \textit{mapping module}, a \textit{planning module}, and a \textit{navigation controller}. The mapping module processes RGB-D inputs and textual prompts to generate an \textit{object-value map}, which integrates detected objects and estimated potential object locations. If a target object is found in the object map, the robot navigates directly to it; otherwise, it relies on the value map to guide exploration. Additionally, the module constructs an obstacle map for navigation constraints and a frontier map to identify unexplored areas. The planning module maintains a belief representation of object locations and employs the POUCT algorithm to simulate and evaluate action sequences, selecting the one with the highest expected reward for execution. The \textit{navigation controller} receives a target location from the planning module and outputs discrete movement commands (\textit{move forward, turn left, turn right}) to guide the robot toward its destination.

\subsection{Mapping}

\subsubsection{Multi-Layer Value Map and Object Map}
Following previous works  VLFM~\cite{vlfm} and Finder~\cite{finder}, we employ a pre-trained BLIP-2~\cite{blip2} vision-language model to compute cosine similarity scores between the robot’s current RGB observation and text prompts corresponding to each target object. At each time step, a cone-shaped confidence mask is generated to represent the camera’s field of view (FOV). The confidence of each pixel is computed as:

\begin{equation}
c(i, j) = \cos^2 \left(\frac{\theta}{\theta_{\text{FOV}} / 2} \times \frac{\pi}{2} \right)
\end{equation}

where \( \theta \) is the angle between the pixel and the optical axis, and \( \theta_{\text{FOV}} \) is the camera's horizontal field of view.

To handle overlapping observations over time, we apply a weighted averaging update for the value map:

\begin{equation}
v_{i,j}^{\text{new}} = \frac{c_{i,j}^{\text{curr}} v_{i,j}^{\text{curr}} + c_{i,j}^{\text{prev}} v_{i,j}^{\text{prev}}}{c_{i,j}^{\text{curr}} + c_{i,j}^{\text{prev}}}
\end{equation}

where \( v_{i,j} \) represents the value at pixel \( (i,j) \), and \( c_{i,j} \) denotes the confidence score. The confidence update is computed as:

\begin{equation}
c_{i,j}^{\text{new}} = \frac{(c_{i,j}^{\text{curr}})^2 + (c_{i,j}^{\text{prev}})^2}{c_{i,j}^{\text{curr}} + c_{i,j}^{\text{prev}}}
\end{equation}

which biases the update towards higher confidence values.

After obtaining the value maps for different target objects, we normalize and aggregate them to form a shared value representation. This step follows a similar approach to previous works~\cite{mos_base,mos_new,mos_system} that assume independence across objects, where the joint belief is computed as the product of individual beliefs.

For object detection, we incorporate Grounding DINO~\cite{dino}, YOLOv7~\cite{yolo}, and Segment Anything Model (SAM)~\cite{sam}. These models enable us to detect, segment, and store all identified objects in an object map throughout the search process.

\subsubsection{Obstacle Map and Frontier Map}
We utilize depth and odometry data to construct a top-down 2D obstacle map, representing regions that the robot has identified as non-traversable. Based on this obstacle map, we determine boundaries between explored and unexplored areas and extract midpoints along these boundaries as potential frontier waypoints. These frontiers guide exploration in unknown environments.

\begin{algorithm}
\caption{\textbf{POUCT-based Planning} $(\mathcal{P}, b_t, d) \to {a}$}
\label{alg:vlm-pouct}
\begin{algorithmic}[1]
\Require $\mathcal{P} = \langle {\mathcal{S}}, {\mathcal{A}}, {\mathcal{O}}, {T}, {O}, R,\gamma \rangle$ 
\Statex \hspace{1.2cm} where
\Statex \hspace{1.2cm}  ${\mathcal{A}} = MoveTo\big\{(x_1^c, y_1^c), \dots, (x_n^c, y_n^c),(x^f, y^f)\big\}$ 
\Statex \hspace{1.2cm} $ b_t = \big\{( (x_1^c, y_1^c): p_1 ), \dots, \big( (x_n^c, y_n^c): p_n \big) \big\}$ 
\Ensure ${a}$: An action in the $\mathcal{A}$ of $\mathcal{P}$
\vspace{0.5em}
\Procedure{Plan}{$b_t$}
    \State $\mathcal{G} \gets$ GenerativeFunction($\mathcal{P}$)
    \State $Q(b_t, {a}) \gets$ POUCT($\mathcal{G}, h_t$)
    \State \Return ${a}$
\EndProcedure
\end{algorithmic}
\end{algorithm}

\subsection{Planning}

We model the planning process within an Object-Oriented POMDP (OO-POMDP) framework~\cite{mos_base}. In our formulation, the state and observation spaces are decomposed with respect to a single \textit{target object}, and multi-object search is achieved via our multi-layer value map using \textit{add} and \textit{norm} operations. Our approach introduces two key modifications: a novel action space formulation and an alternative belief update mechanism for real-world execution.

\subsubsection{Update the \textit{Action Space} and \textit{Belief} with Candidate Points and Selected Frontier}

Let the raw value map be denoted as \( v(x,y) \) over the spatial domain. First, the frontier with the highest value is selected from \( v(x,y) \) as a representative exploratory point \((x^f, y^f)\). Then, to refine the value distribution for targeted search, we apply a decay function:
\begin{equation}
v'(x,y) = \frac{1}{1 + \exp\left(\frac{u(x,y) - \tau}{\kappa}\right)},
\label{eq:decay}
\end{equation}

where \( u(x,y) \) is the update count at location \((x,y)\), and \(\tau\) and \(\kappa\) are constants controlling the decay rate. After thresholding \( v'(x,y) \) to extract high-value regions, we employ DBSCAN~\cite{dbscan} clustering to yield a set of candidate points $(x_i^c, y_i^c)$:
\[
\mathcal{C} = \{(x_i^c, y_i^c) \mid i = 1,\dots,n\}.
\]

The action space is then defined as:
\[
\mathcal{A} = \mathcal{A}_{\text{cand}} \cup \mathcal{A}_{\text{frontier}},
\]
with
\[
\mathcal{A}_{\text{cand}} = \{ \texttt{MoveTo}((x_i^c, y_i^c)) \mid i = 1,\dots,n \},
\]
\[
\mathcal{A}_{\text{frontier}} = \{ \texttt{MoveTo}((x^f, y^f)) \}.
\]
This formulation balances targeted search (via candidate points) with exploratory actions (via the frontier).

The belief over the target object's location is represented as a discrete distribution over candidate points:
\[
b_t = \{\, ((x_i^c, y_i^c): p_i) \mid i = 1,\dots,n \,\},
\]
where
\[
p_i = \frac{v(x_i^c, y_i^c)}{\sum_{j=1}^{n} v'(x_j^c, y_j^c)}.
\]
Here, \(p_i\) denotes the probability associated with the candidate point \((x_i^c, y_i^c)\).

During the POMDP solution process (i.e., in the simulation phase), we still update the belief using the standard Bayesian rule:
\[
b_{t+1}(s') = \eta\, \Pr(o \mid s', a) \sum_{s \in \mathcal{S}} \Pr(s' \mid s, a)\, b_t(s),
\]
with normalization constant \(\eta\). However, in real-world execution, after receiving an observation, we approximate the belief update using a \emph{decayed value map}. This approach circumvents explicit reliance on the observation model while retaining adaptability, and thereby offers a more efficient method for adjusting the belief in practical settings.

The rationale for this approach is that the value map itself already encodes a highly reliable representation of the probability that the object is located at each candidate point. After incorporating detector information into the decayed value map, there is no longer a need to perform extensive POMDP simulations—where possible observations are hypothesized and used in Bayesian updates—to adjust the belief after receiving the real observation. Instead, the decayed value map directly reflects the updated likelihood of the object's location in a simpler, yet still effective, manner.

\subsubsection{POMDP Components}

We model our planning problem as a POMDP defined by the tuple
\[
\langle \mathcal{S}, \mathcal{A}, \mathcal{O}, T, O, R, \gamma \rangle.
\]
The components are defined as follows:

\begin{itemize}
    \item \textbf{State Space (\(\mathcal{S}\))}: An environment state \( s \) is represented as a combination of the robot's state \( s_r \) and the target object's state \( s_t \):
    \[
    s = \{ s_r, s_t \}.
    \]
    The robot state is defined as \( s_r = (x_r, y_r) \) with \( (x_r, y_r) \in \mathbb{R}^2 \), representing its 2D position, while the target object's state is given by \( s_t = (x_t, y_t) \) with \( (x_t, y_t) \in \mathbb{R}^2 \), indicating its position in the environment.

    \item \textbf{Observation Space (\(\mathcal{O}\))}: The robot obtains an observation \( o \) from its RGB-D camera, modeled as a binary indicator of whether detected or not:
    \[
    o \in \{0,1\}.
    \]
    \item \textbf{Action Space (\(\mathcal{A}\))}: Actions are defined as movement commands toward a goal location:
    \[
    a = \texttt{MoveTo}(g)
    \]
    where \( g \) is chosen from candidate points extracted from the value map or the selected frontier.
     \item \textbf{Transition Model (\(T(s,a,s')\))}: The target object is static (i.e., \( s_t' = s_t \)), and the robot state transitions deterministically:
    \[
    s_r' = f(s_r, a),
    \]
    where \( f \) updates the robot's 2D position based on the selected action.
     \item \textbf{Observation Function  (\(O\))}:
    The detection probability is defined as:
   \begin{equation}
    \Pr(o=1 \mid s',a) =
    \begin{cases}
      1, \text{if } d(s_r', s_t) \leq \delta, \\[2mm]
      \exp\Bigl(-\beta\,\bigl(d(s_r', s_t)-\delta\bigr)\Bigr), \\[1mm]
      \quad \text{if } d(s_r', s_t) > \delta.
    \end{cases}
    \end{equation}

    where \( d(s_r', s_t) \) is the Euclidean distance between the robot and the target, \( \delta \) is the detection threshold, and \( \beta > 0 \) controls the decay rate.

    \

    \item \textbf{Reward Function (\(R(s,a)\))}: The reward function is defined as:
    \begin{equation}
    \begin{split}
    R(s,a) =\ & -\lambda_{\text{move}}\, d(s_r, s_r') \\
              & + \lambda_{\text{frontier}}\, \mathbb{I}(a \in \mathcal{A}_{\text{frontier}}) \\
              & + \lambda_{\text{target}}\, \mathbb{I}\Bigl(d(s_r', s_t) \leq \delta\Bigr),
    \end{split}
    \end{equation}
    where \( d(s_r, s_r') \) is the distance traveled by the robot, \(\mathbb{I}(\cdot)\) is the indicator function, and \(\lambda_{\text{move}}, \lambda_{\text{frontier}}, \lambda_{\text{target}} \ge 0\) are weight parameters balancing movement cost, exploratory incentive, and target proximity reward.

    \item \textbf{Discount Factor (\(\gamma\))}: A constant \(\gamma \in (0,1)\) is used to balance immediate and future rewards.
\end{itemize}

To solve the formulated POMDP efficiently, we employ the Partially Observable Upper Confidence Trees (POUCT) algorithm~\cite{pouct}—a Monte Carlo Tree Search-based method. As illustrated in Algorithm~\ref{alg:vlm-pouct}, after the agent executes a step in the real environment, it obtains an updated map from which new candidate points and a frontier are extracted. These are then used to update the action space \( \mathcal{A} \) and the belief \( b_t \). 

Then, it is straightforward to define a generative function
\[
\mathcal{G}(s, a) \rightarrow (s', o, r),
\]
using its transition, observation, and reward functions. Leveraging this generative function, POUCT builds a search tree to simulate transitions, observations, and rewards, thereby planning and selecting the next optimal action.

\section{EXPERIMENTAL RESULTS}
\label{sec:result}

Our experiments consist of two stages: (1) a comprehensive quantitative and qualitative evaluation in a simulation environment, where we compare the success rate and efficiency of our method against recent state-of-the-art (SOTA) approaches and conduct an ablation study to analyze the impact of different components in the OpenGuide system, and (2) a quantitative and qualitative validation on a real robot, where we deploy the agent in a \( 50 \, m^2 \) office environment and a studio-like apartment to search for multiple objects.

\subsection{Experiment Hardware Design}

\begin{figure}
     \centering
     \begin{subfigure}{0.6\textwidth}
         \centering
         \includegraphics[width=\linewidth]{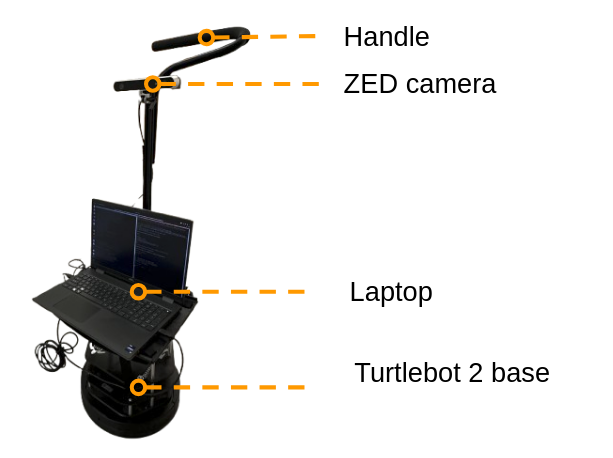}
         \caption{GuideBot Hardware Design}
         \label{fig:hardware}
     \end{subfigure}
     \begin{subfigure}{0.35\textwidth}
         \centering
         \includegraphics[width=\linewidth]{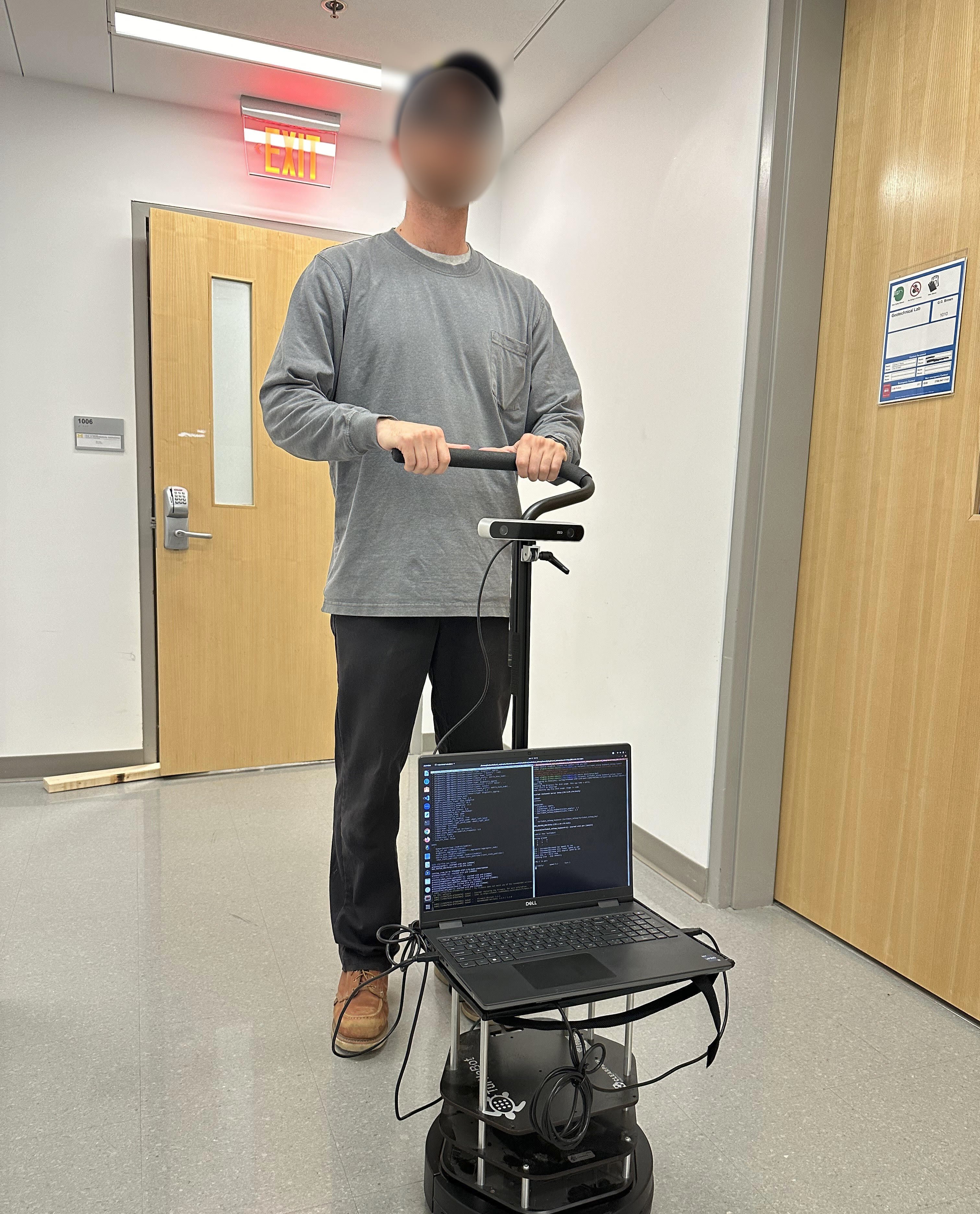}
         \caption{Demonstration of Usage}
         \label{fig:demo}
     \end{subfigure}
     \caption{GuideBot Hardware Design and Demonstration of Usage}
\end{figure}

We developed a GuideBot, a mobile robotic platform designed to assist individuals, particularly those with visual impairments, in navigating complex indoor environments. The system architecture is illustrated in Fig.~\ref{fig:hardware}. Built on a Turtlebot 2 base, GuideBot uses ROS1 Noetic to enable autonomous mobility, perception, and navigation. Mounted on top of the platform is a ZED stereo camera, which provides high-resolution RGB-D data for 3D scene understanding, obstacle avoidance, and semantic perception. A laptop processes the sensor data and runs the ROS-based software stack, handling tasks such as SLAM, planning, and interaction. A distinctive feature of GuideBot is its handlebar, which serves as a physical interface for users to hold onto while being guided through an environment. \ref{fig:demo} illustrates a demonstration of GuideBot guiding a user through an indoor hallway.

Although GuideBot’s overall profile is relatively low, the mounted ZED stereo camera provides a sufficient field of view for detecting both floor-level and tabletop-level obstacles in typical indoor environments. During our experiments, the system demonstrated reliable perception of relevant features due to the wide field of view and depth range of the ZED sensor. Nonetheless, the platform is designed to be modular, and future iterations may incorporate additional or repositioned sensors to further enhance perception in more cluttered or unstructured spaces.

\subsection{Case Study 1: Simulation Environment}

We conduct our simulation experiments using the validation split of the Habitat Matterport3D dataset~\cite{ramakrishnan2021hm3d}. We select five scenes and construct a total of 120 episodes, all of which involve searching for two or three objects. At the beginning of each episode, the robot is initialized at a random location within the environment and provided with a list of target objects. The episode progresses as follows: each time the robot calls \textit{stop}, it indicates that it has found an object. If the distance between the robot and the nearest target object is less than 1\,m at this moment, the object is considered successfully found. The robot then proceeds to search for the remaining objects. The episode terminates when either all target objects have been found or the robot reaches the maximum step limit of 500. 

To evaluate performance, we use two key metrics:

\begin{itemize}
    \item \textbf{Success Rate (SR)}: The percentage of episodes in which the robot successfully finds all target objects.
    \item \textbf{Multi-Object Success weighted by normalized inverse Path Length (MSPL)}: Based on the SPL metric, MSPL is calculated as:
    \begin{equation}
        MSPL = \frac{1}{N} \sum_{i=1}^{N} S_i \frac{l_i}{\max(p_i, l_i)}
    \end{equation}
    where $N$ denotes the total number of episodes, $S_i$ is a binary indicator of success for episode $i$, $l_i$ represents the optimal shortest path length from the start location to all target objects, and $p_i$ denotes the actual path length traversed by the robot.
\end{itemize}

We compare our approach, OpenGuide, against the following baseline methods:
\begin{itemize}
    \item \textbf{Random Walk (Lower Bound)}: A naive strategy where the robot moves randomly in the environment.
    \item \textbf{VLFM~\cite{vlfm}(SOTA for Single Object Search)}: This method combines a VLM-generated value map with frontier-based exploration, selecting the best frontier at each step. In our experiments, the Multi Object Search task is decomposed into a sequence of SOS tasks executed independently.
    \item \textbf{Finder~\cite{finder} (SOTA for Multi object search)}: Built upon VLFM, this method utilizes value maps to compute the \textit{scene-to-object} score and incorporates an additional \textit{object-to-object} score to select the optimal frontier.
\end{itemize}

\begin{table}[t]
    \centering
    \setlength{\tabcolsep}{6pt} 
    \begin{tabular}{l|cc}
        \hline
        Methods & SR↑ & MSPL↑ \\
        \hline
        Random Walk & 0.0\%  & 0.0  \\
        VLFM    & 12.5\%  & 0.075  \\
        Finder  & 28.3\%  & 0.198  \\
        OpenGuide(Ours)    & \textbf{55.0\%}  & \textbf{0.497}  \\
        \hline
    \end{tabular}
    \caption{Overall performance comparison of Random Walk, VLFM~\cite{vlfm}, Finder~\cite{finder}, and Our Method on Multi Object Search, including Success Rate (SR) and Multi-Object Success weighted by normalized inverse Path Length (MSPL).}
    \label{tab:multi_object_search}
\end{table}

\begin{table}[t]
    \centering
    \setlength{\tabcolsep}{4.5pt} 
    \begin{tabular}{l|cc}
        \hline
        Methods & SR↑ & MSPL↑ \\
        \hline
        OpenGuide w/o POMDP & 27.5\%  & 0.154  \\
        OpenGuide w/o value decay & 45.0\%  & 0.410 \\
        OpenGuide (Full) & \textbf{55.0\%}  & \textbf{0.497}  \\
        \hline
    \end{tabular}
    \caption{Ablation study on OpenGuide, evaluating the impact of different components, including POMDP and value decay introduced in Eq.~\eqref{eq:decay}. 
    }
    \label{tab:ablation_study}
\end{table}

\begin{figure}[t!]
    \centering
    \includegraphics[width = \textwidth]{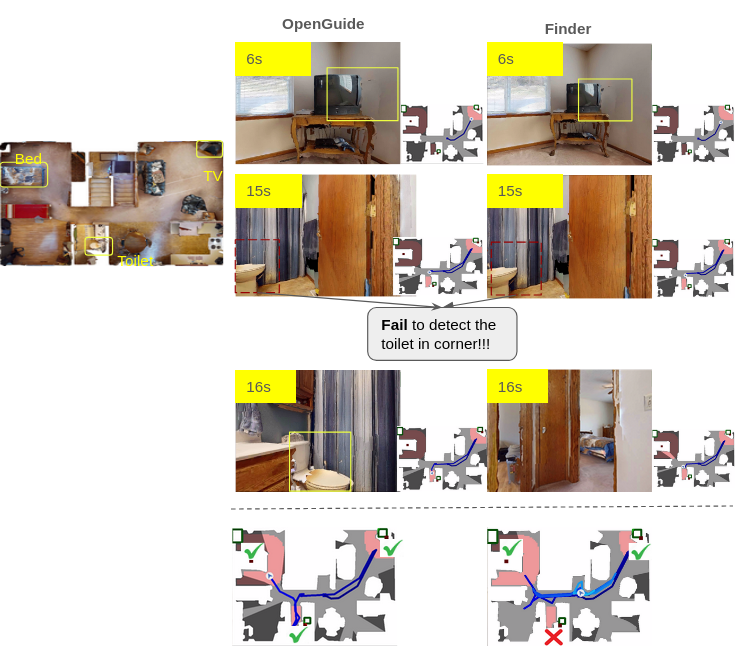}
    \caption{Qualitative comparison between OpenGuide and Finder~\cite{finder} in simulation.}
    \label{fig:sim_qua}
\end{figure}

The experimental results are summarized in Table~\ref{tab:multi_object_search}. VLFM achieves only a 12.5\% success rate (SR) with an MSPL of 0.075, which reflects its limitation of tracking only one object at a time. Finder improves on this by employing a multi-layer value map to simultaneously track multiple objects, resulting in an enhanced SR of 28.3\% and an MSPL of 0.198. Our method further builds on Finder’s strengths by integrating POMDP-based planning into the framework. This additional planning mechanism enables adaptive exploration and more informed decision-making, leading to a significant performance boost with an SR of 55.0\% and an MSPL of 0.497. Thus, OpenGuide nearly doubles the success rate and more than doubles the MSPL compared to Finder, demonstrating its superior accuracy and efficiency in multi-object search tasks.

The key advantage of OpenGuide over Finder lies in scenarios where objects are occluded or partially visible. As illustrated in Fig.~\ref{fig:sim_qua}, the front row is the demonstration of our proposed OpenGuide method and the bottom one is the Finder baseline. in some episodes, due to occlusions and viewing angles, a toilet located in the corner was not detected in the already-explored region. Finder, relying on its frontier-based strategy, abandoned this area and navigated toward a different frontier, ultimately never revisiting the corner and failing to find the toilet. In contrast, OpenGuide recognized the region's high value and continued further exploration, eventually obtaining an effective viewpoint to detect the object. This ability to revisit and refine exploration based on value map information provides OpenGuide with a distinct advantage over Finder in complex search scenarios.We did not compare with VLFM and Random Walk here because their performance is significantly weaker than OpenGuide and Finder.

In addition, we conduct an ablation study to analyze the impact of key components in OpenGuide, specifically the absence of POMDP-based planning and the absence of value decay. The results in Table~\ref{tab:ablation_study} indicate that removing POMDP leads to much worse performance. Without POMDP, the agent directly selects the highest-value frontier or candidate point at each step, resulting in an overly greedy strategy. This approach fails to account for the navigation effort required to reach each point and does not balance the trade-off between exploring new frontiers and exploiting high-value candidate points, leading to suboptimal overall performance.

For the case without value decay in Table~\ref{tab:ablation_study}, we observe a drop in success rate. Further analysis reveals that in some episodes, certain regions maintained high-value scores despite not containing the actual target objects. As a result, the robot spent excessive steps repeatedly exploring these high-value areas without making progress, sometimes getting trapped and ultimately failing the task. This highlights the importance of value decay in preventing the robot from fixating on misleading high-value regions and ensuring more effective search behavior.

\subsection{Case Study 2: Home Environment}

\begin{table}[ht]
\centering
\footnotesize
\begin{tabular}{>{\RaggedRight\arraybackslash}>{\itshape}p{5cm}c|c|c}
\toprule
Instructions & Inferred Object Type& Success/Total & Average Steps \\
\toprule
Find a bottle and a potted plant. I want to water the plant. &bottle, potted plant& 1/3 & 104 \\
Please locate the microwave and a bowl so I can warm up my food. &microwave, bowl& 3/3 & 70 \\
Can you check where the refrigerator, my bed, and a bowl are? I want to prepare for a rest after eating. & refrigerator, bed, bowl& 1/3 & 95.3 \\
Find my laptop and couch so I can get to work. &couch, laptop& 2/3 & 34.7\\
Find the fork, bowl, and spoon so I can have my dinner. & fork, bowl, spoon& 0/3 & 200 \\
Please find the toothbrush and the sink. I need to brush my teeth. & toothbrush, sink& 2/3 & 138.3 \\ \hline
\textbf{Average} &  & 9/18 & 107.6\\
\bottomrule
\end{tabular}
\caption{Quantitative Result of OpenGuide in home environment}
\label{tab:quantitative 2}
\end{table}


\begin{figure}
    \centering
    \includegraphics[width = 0.7\textwidth]{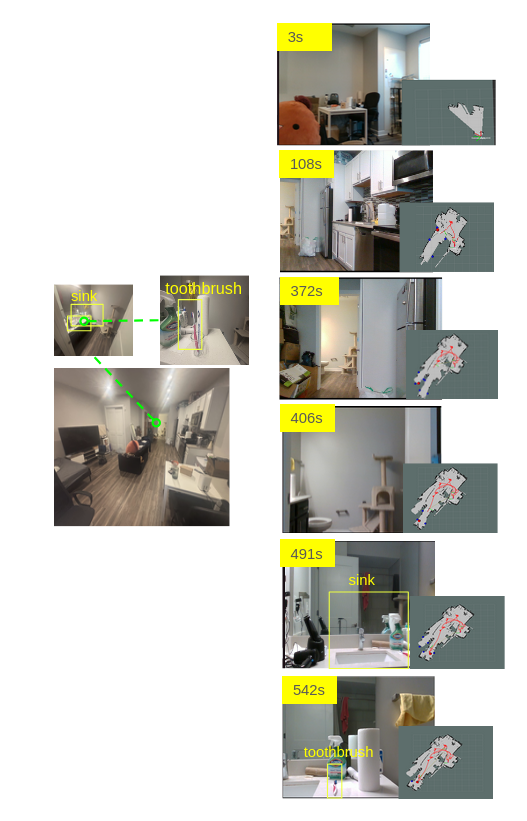}
    \caption{Qualitative Results of OpenGuide in home environment.}
    \label{fig:real_qual_apartment}
\end{figure}

We conduct a real-world case study in a home environment to prove the simulation result. The object layout of the home environment is shown in the left side of Fig.~\ref{fig:real_qual_apartment}. The environment is a 470 square foot studio-like student apartment with objects clustered together. In the home environment, we have conducted six sets of multi-object searching tasks, and each task contains two to three objects. The quantitative result of the case study inside the home environment is shown in Table~\ref{tab:quantitative 2}. The system achieved an average object search success rate of approximately 50\%, which closely aligns with the performance observed in the simulated environment. This consistency highlights the robustness and generalizability of OpenGuide across environments with varying spatial layouts and visual characteristics. The average number of navigation steps was 107.6—higher than in simulation—which is expected given the increased difficulty posed by navigation errors and densely packed objects.

The qualitative result is illustrated in Fig.~\ref{fig:real_qual_apartment}. In this example, the robot is tasked with finding a "toothbrush" and a "sink" located in the bathroom, starting from an initial position in the living room. At the beginning, the robot has no visual cues related to either target object, prompting it to initiate exploratory behavior by selecting frontier points planned through a POMDP-based policy. As the robot navigates through the apartment, it begins exploring the bathroom at around 372 seconds. Utilizing segmentation and grounding results from the Vision-Language Model (VLM), the robot successfully identifies the sink at 491 seconds and the toothbrush at 542 seconds.

\subsection{Case Study 3: Office Environment}

\begin{table}[t]
\centering
\footnotesize
\begin{tabular}{>{\RaggedRight\arraybackslash}>{\itshape}p{5cm}c|c|c}
\toprule
Instructions & Inferred Object Type& Success/Total & Average Steps \\
\toprule
Please search the break room for a book and a reusable bottle I left. & bottle, book& 1/3 & 135 \\
Can you take me to the refrigerator and help me find a bowl for lunch? & refrigerator, bowl& 2/3 & 24 \\
Raining outside, can you find my umbrella and the backpack I brought this morning? & umbrella, backpack& 2/3 & 98.3 \\
I want to sit down and read—can you find a chair and a book? & chair, book& 2/3 & 64 \\
I’m going to have lunch. Please fetch a bowl and a cup, and find me a couch to sit on. & cup, bowl, couch& 2/3 & 94.3\\
Help me find a laptop and keyboard, then bring a chair so I can get started. & keyboard, laptop, chair& 1/3 & 143.7 \\
Please bring me the spray bottle and guide me to the potted plant that needs watering. & potted plant,  bottle & 3/3 & 36.3 \\
Can you guide me to the couch in front of the TV? I want to take a break. & tv, couch& 2/3 & 18.7\\
I left an apple and a banana on my desk. & apple,banana& 0/3 & 151.7 \\
Can you bring me my laptop and guide me to a couch where I can work comfortably? & couch,laptop& 2/3 & 14.3 \\ \hline
\textbf{Average} &  & 17/30 & 80.6\\
\bottomrule
\end{tabular}
\caption{Quantitative Result of OpenGuide in office environment}
\label{tab:quantitative 3}
\end{table}

\begin{figure}
    \centering
    \includegraphics[width = 0.7\textwidth]{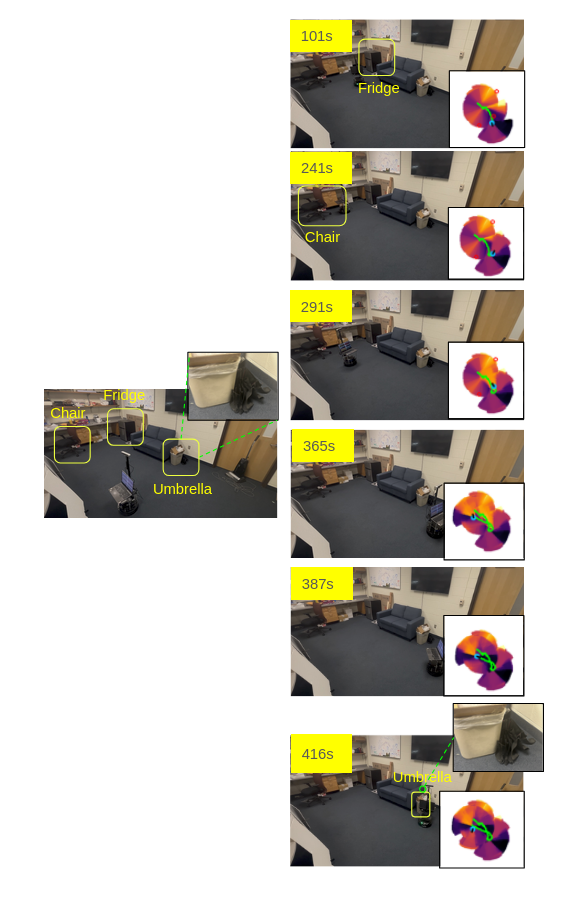}
    \caption{Qualitative results of OpenGuide in an office environment}
    \label{fig:real_qual}
\end{figure}

Besides the simulation and real robot experiment in a household environment, we also conducted an experiment in an office environment. We first task the robot with searching for objects in a cluttered \(50\,m^2\) office environment, including easily detectable objects (a chair and a fridge) and a more challenging object (an umbrella occluded by a trash can). We let the robot find ten sets of objects (each one contains two or three objects), and for each set, we test it three times. 

The quantitative result is shown in Table~\ref{tab:quantitative 3}. The table presents a qualitative evaluation of OpenGuide, showcasing its performance across 10 representative natural language instructions in an office-like environment. Out of a total of 30 object instances across all instructions, the system successfully retrieved 17, resulting in a 56.7\% success rate. Notably, the system demonstrated perfect performance (3/3) in retrieving objects with strong spatial or functional associations (e.g., "potted plant" and "spray bottle"). The average number of steps required to complete each task was 80.6. The number of navigation steps is strongly influenced by the spatial distribution of target objects and the number of objects to find within one set. Instructions involving objects located close to the robot's initial position or clustered in the same area led to lower step counts. In contrast, when the objects were placed in distant or separate rooms, the system required substantially more steps. 

As for the qualitative result shown in Fig.~\ref{fig:real_qual}, initially the fridge and chair were easily detected due to their large size and clear features. In particular, the fridge was found at 101s and the chair at 241s, while the umbrella remained hidden because of unfavorable viewpoints. Once the entire environment had been explored—and from 365s onward, the value map no longer showed any frontiers—the robot continuously replanned its trajectory based on the updated value map information. This iterative replanning eventually led to an effective viewpoint from which the occluded umbrella was detected. 

This experiment demonstrates the strong recovery and replanning capability of OpenGuide in real-world scenarios, especially when the environment has been fully explored, yet the target object remains undetected.

\subsection{Discussion}

\begin{figure}[t]
    \centering
    \includegraphics[width = 1\textwidth]{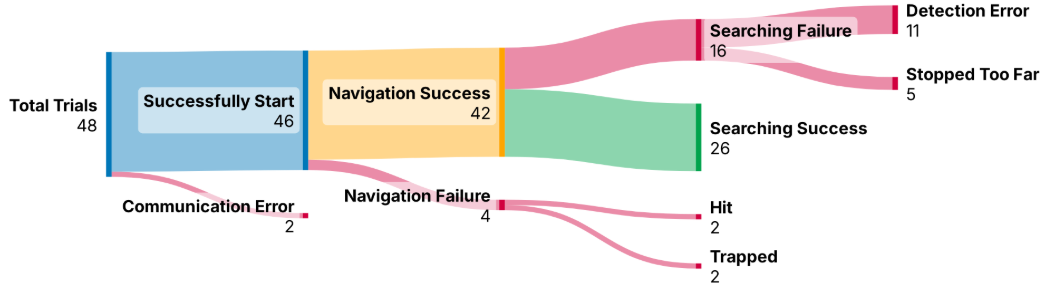}
    \caption{Summary of success and failure cases of real world experiment}
    \label{fig:summary}
\end{figure}

We conducted a total of 48 real-world multi-object search trials, comprising 18 in a home environment and 30 in an office environment. The breakdown of task outcomes is visualized in Fig.~\ref{fig:summary}. These trials reveal key insights into the end-to-end performance of our system and highlight critical areas for improvement.

Out of the 48 trials, 46 were successfully initiated, with 2 failures due to communication errors such as system disconnects or command timeouts. Among the initiated trials, 42 reached the designated search regions, demonstrating robust navigation performance even in cluttered or constrained indoor environments. However, 4 trials failed during navigation due to either SLAM-related collisions ("hit") or the robot getting trapped in narrow or occluded spaces. These navigation failures underscore the gap between real-world and simulated environments, where physical constraints, sensor noise, and unexpected obstacles introduce additional challenges that are often abstracted away in simulation.

The most significant performance bottleneck emerged during the object searching phase. While 26 out of the 42 navigation-successful trials successfully identified all target objects, the remaining 16 failed due to perception-related limitations. Specifically, 11 failures were attributed to detection errors—where the robot either misidentified or completely missed the target object. An additional 5 failures occurred because the robot stopped too far from the target location, indicating that the spatial termination of navigation needs to be better aligned with the effective range of perception models.

Overall, our system achieved a 54.2\% end-to-end task success rate (26/48), closely mirroring the results observed in simulation and confirming the system’s consistency across diverse real-world settings.

\section{Conclusion}
\label{sec:conclusion}

In this work, we present OpenGuide, a novel open-vocabulary assistive robotic framework that enables reliable and efficient multi-object retrieval in previously unseen or dynamically evolving indoor environments, specifically tailored for individuals who are blind or visually impaired. OpenGuide integrates VLMs to semantically interpret user-specified object queries and formulates the object search task as a POMDP to reason under uncertainty and make informed navigation decisions. To improve search efficiency, we incorporate frontier-based exploration with language-driven priors, allowing the system to dynamically prioritize regions of interest while minimizing travel cost. Through extensive real-world and simulated evaluations across three diverse case studies, we demonstrate that OpenGuide significantly outperforms existing baselines in both retrieval accuracy and navigation efficiency.

We evaluated OpenGuide across simulation and real-world settings, including both home and office environments. The system consistently achieved strong task success rates—55.0\% in simulation, 50.0\% in a studio-style apartment, and 56.7\% in a cluttered office—demonstrating its generalizability and robustness across domains. The robot successfully handled diverse object categories and instructions grounded in natural language, and its replanning mechanism showed strong recovery from missed detections, outperforming prior methods. 

Despite these promising results, several limitations remain. Most notably, the system's object detection pipeline still struggles with small, occluded, or visually ambiguous items, leading to detection failures. The reliance on static vision-language grounding also limits adaptability in dynamic or heavily cluttered scenes where object appearances vary significantly from pretrained priors. In future work, we aim to enhance the robustness of the visual perception module by trying different vision language backbones~\cite{ren2024grounded}, improve navigation policies with an adaptive obstacle avoidance policy~\cite{xu2024}, and explore dialog-based interaction to allow users to guide or correct the robot in real time. These improvements would further strengthen OpenGuide’s applicability in real-world assistive scenarios in the built environment and move toward more human-centered and collaborative robot assistance.

\section{Data Availability Statement}
\label{sec:data}
Some data, models, or code that support the findings of this study are available from the corresponding author upon reasonable request.
\section{Acknowledgements}
\label{sec:acknowledge}
The work presented in this paper was supported financially by the United States National Science Foundation (NSF) via Award\# SCC-IRG 2124857. The support of the NSF is gratefully acknowledged.
\bibliographystyle{elsarticle-num}
\bibliography{ascexmpl-new}



\end{document}